\documentclass[11pt]{article}
\usepackage{acl}

\usepackage{times}

\usepackage[T1]{fontenc}

\usepackage[utf8]{inputenc}

\usepackage{amsmath,amssymb,amsfonts}
\usepackage{graphicx}
\usepackage{booktabs}
\usepackage{multirow}

\usepackage{float}

\usepackage{xcolor}
\usepackage{microtype}
\usepackage{enumitem}
\usepackage{caption}
\usepackage{tikz}
\usetikzlibrary{arrows.meta,positioning,fit,backgrounds}

\title{SpanCalib-VLM: Calibrated Hallucination Span Detection in Vision-Language Models}

\author{
  \textbf{Amanuel Gizachew Abebe} \vspace{5pt} \\
  Shaggar Institute of Technology \\
  Shaggar City, Ethiopia \\
  \texttt{\small amanuel.g.abebe1@gmail.com} \\
  \And
  \textbf{Yasmin Moslem} \vspace{5pt}\\
  Trinity College Dublin \\
  Dublin, Ireland \\
  \texttt{\small yasmin.moslem@adaptcentre.ie} 
}

\date{}

\begin{document}

\maketitle

\begin{abstract}
Detecting hallucinations in Large Vision-Language Models (LVLMs) requires both accurate span localization and well-calibrated confidence scores. Fine-tuned generative VLMs excel at identifying hallucinated text spans but suffer from overconfidence and high inference latency. Discriminative sequence taggers offer deterministic speed and superior calibration but exhibit conservative span recall. We present \textbf{SpanCalib-VLM}, a hybrid dual-system for the SHROOM-Visions Shared Task that combines a multimodal sequence tagger, consisting of \textit{XLM-RoBERTa-Large} fused with a \textit{SigLIP} vision encoder via cross-attention, with our fine-tuned generative VLM (\textit{Qwen3.5-4B-SHROOM-SFT}). Through a \emph{Union-Calibrated Fusion} strategy, candidate spans from the generative model are re-scored with calibrated probabilities from the sequence tagger. On the SHROOM-Visions English validation split, our ensemble achieves a Pearson calibration correlation of \textbf{0.41} and an overall IoU of \textbf{0.39} (with a clean-response IoU of \textbf{0.91} and overall detection accuracy of \textbf{70.7\%}). On the official blind test set, SpanCalib-VLM achieves a Pearson correlation of \textbf{0.32} and an IoU of \textbf{0.25} on English, substantially outperforming the shared task baseline (0.15 on both metrics) and demonstrating consistent gains across all four languages. We make our model weights and code publicly available.
\end{abstract}

%% ==========================
\section{Introduction}
%% ==========================

Large Vision-Language Models (LVLMs) such as GPT-4V, Qwen-VL \citep{bai2023qwenvl}, and MiniCPM-V \cite{yao2024minicpm,yu2025minicpmv45} have made remarkable progress in visual question answering and multimodal reasoning. However, they regularly produce \emph{visual hallucinations}, i.e. text that misdescribes scenes, invents objects, or mischaracterizes attributes~\cite{li2023evaluating,liu2024survey}.

The SHROOM-Visions shared task~\cite{mickus2024shroom,vazquez2025mushroom} targets fine-grained hallucination detection, requiring systems to (i)~identify character-level span boundaries of hallucinated segments and (ii)~provide well-calibrated probability scores evaluated via Pearson correlation.

Existing approaches fall into two paradigms. \emph{Generative VLM fine-tuning} trains autoregressive models to output structured span annotations; these achieve high span IoU but suffer from overconfidence and high latency (up to 23\,s per sample with chain-of-thought). \emph{Discriminative sequence tagging} applies token-level classifiers that run in a single forward pass with well-calibrated outputs, but produce conservative span boundaries.

SpanCalib-VLM bridges this gap through four contributions:
\begin{enumerate}[noitemsep,leftmargin=*]
    \item A \textbf{multimodal sequence tagger} combining \textit{XLM-RoBERTa-Large} with \textit{SigLIP} vision features via cross-attention, equipped with multi-task heads for span detection, probability calibration, and error categorization.
    \item \textbf{Union-Calibrated Fusion}, an ensemble algorithm that uses generative VLM spans as candidate proposals and re-calibrates them with the tagger's probability distribution.
    \item Competitive results on SHROOM-Visions: achieving \textbf{Pearson~0.413} and \textbf{IoU~0.391} on the English validation split, and \textbf{Pearson~0.3212} and \textbf{IoU~0.2549} on the official blind test set, outperforming the task baseline (0.1549).
    \item Public release of model weights, code, and evaluation artifacts.\footnote{\scriptsize\url{https://github.com/Aman-byte1/Hallucination-Detection-in-LVLMs}}
\end{enumerate}

%% ==========================
\section{Related Work}
%% ==========================

\paragraph{Hallucination in VLMs.}
Benchmarks such as POPE~\cite{li2023evaluating}, CHAIR~\cite{rohrbach2018object}, and MME~\cite{fu2025mme} evaluate hallucination at the sentence or object level. The SHROOM series~\cite{mickus2024shroom,vazquez2025mushroom} raises the bar by requiring token-level span identification paired with probability calibration, a joint localization-and-uncertainty problem.

\paragraph{Calibration.}
Modern neural networks and pretrained language models exhibit severe miscalibration and overconfidence, particularly under domain or distribution shift~\cite{desai-durrett-2020-calibration, ovadia2019trustmodelsuncertaintyevaluating, guo2017calibration, kadavath2022languagemodelsmostlyknow}. Standard post-hoc calibration methods like temperature scaling and Platt scaling adjust global logit temperature but fail to capture fine-grained, token-dependent span uncertainty. Rather than relying on uncalibrated autoregressive generation probabilities, our approach equips the discriminative tagger with a dedicated MSE regression head to directly supervise token-level confidence against continuous human consensus ratios.

\paragraph{Hybrid architectures.}
Dual-process cognitive theory distinguishes fast, intuitive pattern recognition (System~1) from slow, deliberative reasoning (System~2), a paradigm increasingly operationalized to combine complementary strengths in artificial intelligence~\cite{booch2020thinkingfastslowai,li202512surveyreasoning}. In sequence processing and multimodal detection, generative VLMs excel at open-ended contextual reasoning and span proposal generation (System~2), but incur high computational latency and poor probability calibration. Conversely, discriminative sequence taggers evaluate representations in a single deterministic forward pass (System~1), providing low latency and well-calibrated token probabilities. SpanCalib-VLM operationalizes this hybrid synergy by pairing a discriminative tagger for probability calibration with a generative VLM for candidate span proposal generation.

%% ==========================
\section{Methodology}
%% ==========================

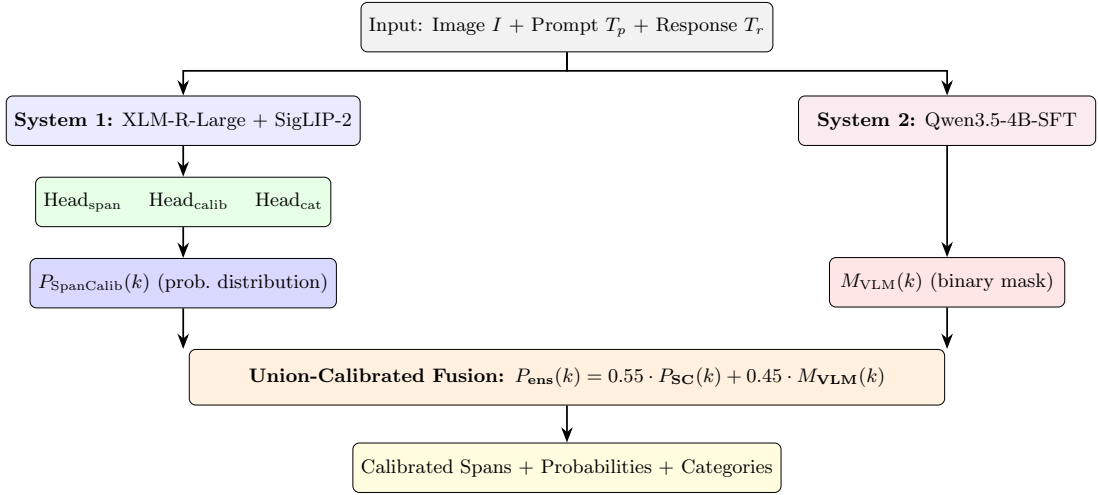
\begin{figure*}[t]
\centering
\resizebox{0.90\textwidth}{!}{%
\begin{tikzpicture}[
    node distance=0.6cm and 0.8cm,
    box/.style={draw, rounded corners=3pt, minimum height=0.8cm,
                text centered, font=\small},
    sysbox/.style={box, fill=blue!8, minimum width=4.8cm},
    headbox/.style={box, fill=green!10, minimum width=4.2cm},
    fusebox/.style={box, fill=orange!12, minimum width=12.2cm, minimum height=0.9cm, font=\small\bfseries},
    arr/.style={-{Stealth[length=2.5mm]}, thick},
]
% Input
\node[box, fill=gray!10, minimum width=5cm] (input) {Input: Image $I$ + Prompt $T_p$ + Response $T_r$};

% System 1
\node[sysbox, below left=0.7cm and 0.0cm of input] (s1) {\textbf{System 1:} XLM-R-Large + SigLIP-2};
\node[headbox, below=0.5cm of s1] (heads) {$\text{Head}_{\text{span}}$ \quad $\text{Head}_{\text{calib}}$ \quad $\text{Head}_{\text{cat}}$};
\node[box, fill=blue!15, below=0.5cm of heads, minimum width=3.8cm] (probs) {$P_{\text{SpanCalib}}(k)$ (prob.\ distribution)};

% System 2
\node[sysbox, fill=purple!8, below right=0.7cm and 0.4cm of input] (s2) {\textbf{System 2:} Qwen3.5-4B-SFT};
\node[box, fill=red!10, below=1.8cm of s2, minimum width=3.8cm] (mask) {$M_{\text{VLM}}(k)$ (binary mask)};

% Fusion (centered under input)
\node[fusebox, below=4.8cm of input] (fusion) {Union-Calibrated Fusion: $P_{\text{ens}}(k) = 0.55 \cdot P_{\text{SC}}(k) + 0.45 \cdot M_{\text{VLM}}(k)$};

% Output
\node[box, fill=yellow!15, below=0.6cm of fusion, minimum width=6cm] (out) {Calibrated Spans + Probabilities + Categories};

% Arrows
\draw[arr] (input.south) -- ++(0,-0.3) -| (s1.north);
\draw[arr] (input.south) -- ++(0,-0.3) -| (s2.north);
\draw[arr] (s1) -- (heads);
\draw[arr] (heads) -- (probs);
\draw[arr] (s2) -- (mask);
\draw[arr] (probs.south) -- (probs.south |- fusion.north);
\draw[arr] (mask.south) -- (mask.south |- fusion.north);
\draw[arr] (fusion) -- (out);
\end{tikzpicture}%
}
\caption{SpanCalib-VLM architecture. System~1 (discriminative tagger) produces calibrated token probabilities; System~2 (generative VLM) proposes candidate spans. Union-Calibrated Fusion combines both into final predictions.}
\label{fig:arch}
\end{figure*}

In this section we elaborate on our approach. Figure~\ref{fig:arch} illustrates the overall architecture of our SpanCalib-VLM system.

% Figure

\subsection{Task Formalization}

Given an image~$I$, prompt~$T_p$, and LVLM response~$T_r$ of length~$L$, the gold standard provides character-level annotations $\mathcal{G} = \{(s_i, e_i, c_i, p_i)\}_{i=1}^{M}$ where $s_i,e_i \in [0,L]$ are character offsets, $c_i$ is a hallucination category, and $p_i \in [0,1]$ is a target probability representing human annotator consensus confidence. Systems must predict matching character-level span boundaries paired with well-calibrated continuous probability scores.

\subsection{System 1: Multimodal Sequence Tagger}

\paragraph{Text encoder.}
We use \textit{XLM-RoBERTa-Large} \citep{conneau2020xlmroberta} (24 layers, $d_h\!=\!1024$). The input concatenates the prompt and response as ``\textit{Question:} $T_p$ \textit{\textbackslash nResponse:} $T_r$'' and is tokenized to a maximum of $N\!=\!512$ tokens, yielding hidden states $H_{\text{text}} \in \mathbb{R}^{N \times d_h}$.

\paragraph{Vision encoder \& cross-attention fusion.}
Image features are extracted with \textit{SigLIP}\footnote{\scriptsize\url{https://hf.co/google/siglip-base-patch16-224}} \citep{zhai2023siglip}, producing patch embeddings $V \in \mathbb{R}^{196 \times 768}$, which are linearly projected into the text hidden space ($V_{\text{proj}} \in \mathbb{R}^{196 \times d_h}$). Cross-attention then fuses visual context into language representations:

\begin{small}
\begin{align}
    H_{\text{fused}} &= \mathrm{LayerNorm}\!\left(H_{\text{text}} + \mathrm{Softmax}\!\left(\tfrac{QK^\top}{\sqrt{d_h}}\right) U\right)
\end{align}
\end{small}

\paragraph{Multi-task heads.}
Three parallel heads operate on each token~$h_i$:
\begin{enumerate}[noitemsep,leftmargin=*]
    \item \textbf{Span classifier} ($\mathrm{Head}_{\text{span}}$): 2-layer MLP $\rightarrow \sigma(\cdot)$, predicting binary hallucination probability $\hat{y}_i \in [0,1]$.
    \item \textbf{Probability calibrator} ($\mathrm{Head}_{\text{calib}}$): 2-layer MLP $\rightarrow \sigma(\cdot)$, predicting continuous calibration score $\hat{p}_i \in [0,1]$.
    \item \textbf{Category classifier} ($\mathrm{Head}_{\text{cat}}$): 2-layer MLP $\rightarrow \mathrm{softmax}(\cdot)$, predicting error type among five categories.
\end{enumerate}

\paragraph{Training objective.}
The composite multi-task loss is:
\begin{equation}
    \mathcal{L} = \mathcal{L}_{\text{BCE}}(\hat{y}, y) + \lambda_1 \mathcal{L}_{\text{MSE}}(\hat{p}, p) + \lambda_2 \mathcal{L}_{\text{CE}}(\hat{\mathbf{c}}, \mathbf{c})
    \label{eq:loss}
\end{equation}
with $\lambda_1\!=\!1.0$ and $\lambda_2\!=\!0.5$. The MSE term directly supervises probability calibration against annotator agreement scores $p$. The auxiliary classification loss $\mathcal{L}_{\text{CE}}$ acts as a multi-task regularizer, forcing shared encoder representations to capture fine-grained semantic category boundaries.

\subsection{System 2: Generative VLM}

We fine-tune \textit{Qwen3.5-4B} \citep{yang2025qwen3,qwen3.5} via supervised fine-tuning (SFT) on SHROOM-Visions training data \citep{mickus2026shroom} to produce structured JSON spans. From System\,2's output, candidate spans $\mathcal{S}_{\text{VLM}}$ are converted into a binary character mask $M_{\text{VLM}}(k) \in \{0,1\}$.

\subsection{Union-Calibrated Fusion}

Neither system alone optimizes both localization and uncertainty estimation: System\,2 provides high span recall through deliberative generation but outputs uncalibrated scores, while System\,1 provides well-calibrated probabilities but conservative span boundaries. Our fusion algorithm operates as follows:
\begin{enumerate}[noitemsep,leftmargin=*]
    \item Extract candidate span boundaries $\mathcal{S}_{\text{VLM}}$ from System\,2.
    \item Map System\,1's token probabilities to a character-level array $P_{\text{SC}}(k)$ via subtoken offset alignment across the full response sequence.
    \item Compute the probability-guided ensemble score:
    \begin{equation}
        P_{\text{ens}}(k) = w_1 \cdot P_{\text{SC}}(k) + w_2 \cdot M_{\text{VLM}}(k)
    \end{equation}
    with $w_1\!=\!0.55$, $w_2\!=\!0.45$. The weights $w_1$ and $w_2$ were selected via hyperparameter grid search on the validation set: assigning slightly higher weight to System\,1 ($w_1\!=\!0.55$) anchors score magnitudes to its superior calibration baseline (Pearson 0.369 vs. 0.285), while $w_2\!=\!0.45$ allows System\,2's binary mask $M_{\text{VLM}}(k) \in \{0,1\}$ to act as a spatial proposal weight for candidate span positions.
\end{enumerate}

%% ==========================
\section{Experimental Setup}
%% ==========================

\paragraph{Data.}
We fine-tune models on 13,594 training samples from a multilingual dataset(with 1,508 held out for validation) in EN, FR, IT, and ZH (cf.~Table~\ref{tab:dataset_stats}), evaluating main baselines on the English validation split and cross-lingual performance across all four languages on validation (cf.~Table~\ref{tab:multilingual_breakdown}) and test sets (cf.~Table~\ref{tab:official_submissions}).

\paragraph{Training.}
SpanCalib-VLM is trained for 5~epochs on NVIDIA A40 GPUs with AdamW ($\eta\!=\!1.5{\times}10^{-5}$), linear warmup (10\%), batch size 16, and max sequence length 512. Table~\ref{tab:hyperparams} summarizes key hyperparameters.

\paragraph{Training Dynamics}

Table~\ref{tab:epochs} tracks validation metrics across epochs. Pearson correlation peaks at epoch\,2 (0.3688); subsequent epochs exhibit overfitting, with training loss continuing to fall while validation loss rises. The best-checkpoint strategy based on validation Pearson successfully preserves the optimal state.

\begin{table}[H]
\centering
\caption{Training configuration.}
\label{tab:hyperparams}
\small
\begin{tabular}{@{}ll@{}}
\toprule
\textbf{Parameter} & \textbf{Value} \\
\midrule
Text backbone & \textit{xlm-roberta-large} (355\,M) \\
Vision encoder & \textit{siglip-base-patch16-224} (86\,M) \\
Optimizer & AdamW ($\beta_1\!=\!0.9$, $\beta_2\!=\!0.999$) \\
Learning rate & $1.5 \times 10^{-5}$ \\
LR schedule & Linear decay + 10\% warmup \\
Weight decay & 0.01 \\
Batch size & 16 \\
Epochs & 5 (best at epoch 2) \\
Max seq.\ length & 512 tokens \\
Loss weights & $\lambda_1\!=\!1.0$, $\lambda_2\!=\!0.5$ \\
\bottomrule
\end{tabular}
\end{table}

\begin{table}[htbp]
\caption{Training configuration for Qwen3.5-4B (Generative VLM).}
\label{tab:qwen_hyperparams}
\small
\begin{tabular}{@{}ll@{}}
\toprule
\textbf{Parameter} & \textbf{Value} \\
\midrule
Base model & \textit{Qwen3.5-4B} (4.4\,B) \\
Finetuning method & LoRA ($r=16$, $\alpha=16$, all-linear) \\
Precision & BF16 \\
Optimizer & 8-bit AdamW \\
Learning rate & $2.0 \times 10^{-4}$ \\
LR schedule & Linear decay + 5\% warmup \\
Batch size & 8 (batch 2 $\times$ 4 grad accum) \\
Epochs & 3 \\
Max seq.\ length & 2048 tokens \\
Modules saved & \textit{lm\_head}, \textit{embed\_tokens} \\
\bottomrule
\end{tabular}
\end{table}

%% ==========================
\section{Results}
%% ==========================

\subsection{Main Results}

Table~\ref{tab:main} compares standalone baselines and our full system on the English validation split ($N\!=\!379$), which served as the development set for checkpoint selection, fusion weights, and decision thresholds. On this split, SpanCalib-VLM achieves a Pearson correlation of \textbf{0.413} and an overall IoU of \textbf{0.391}, with a clean-response IoU of \textbf{0.913} and detection accuracy of \textbf{70.7\%}. We emphasize that these headline figures reflect validation performance under tuned hyperparameters; on the official blind test set (Table~\ref{tab:official_submissions}), performance is \textbf{0.3212} Pearson and \textbf{0.2549} IoU. While this drop from validation (particularly in IoU, $0.391 \rightarrow 0.2549$) highlights the challenge of generalizing thresholded span boundaries to unseen test distributions with annotator variability, SpanCalib-VLM still achieves a $+64.5\%$ relative improvement in IoU and more than doubles the Pearson correlation over the task baseline ($0.1549$).

\begin{table*}[t]
\centering
\caption{Performance on the SHROOM-Visions English validation split. Our full hybrid system (\textbf{SpanCalib-VLM}) achieves the highest Pearson calibration correlation and overall IoU, outperforming standalone baselines and individual subsystems across all metrics at optimal decision thresholds.}
\label{tab:main}
\resizebox{\textwidth}{!}{%

\begin{tabular}{@{}lccccc@{}}
\toprule
\textbf{System} & \textbf{Pearson} & \textbf{Overall IoU} & \textbf{Halluc.\ IoU} & \textbf{Clean IoU} & \textbf{Det.\ Acc.} \\
\midrule
\multicolumn{6}{l}{\textit{Baselines}} \\
BLIP baseline & 0.124 & 0.210 & 0.082 & 0.650 & 42.1\% \\
MiniCPM-V (fine-tuned) & 0.245 & 0.298 & 0.121 & 0.740 & 51.5\% \\
\addlinespace
\multicolumn{6}{l}{\textit{Standalone Subsystems (Ours)}} \\
System\,1: Tagger (Text-only, XLM-R Base) & 0.342 & 0.295 & 0.101 & 0.882 & 52.1\% \\
System\,1: Tagger (Text-only, XLM-R Large) & 0.368 & 0.312 & 0.115 & 0.901 & 56.4\% \\
System\,1: Tagger (Multimodal, XLM-R + SigLIP) & 0.369 & 0.326 & 0.155 & 0.903 & 58.0\% \\
System\,2: Generative VLM (Qwen3.5-4B SFT) & 0.285 & 0.375 & 0.182 & 0.885 & 68.2\% \\
\addlinespace
\multicolumn{6}{l}{\textit{Full Hybrid System}} \\
\textbf{SpanCalib-VLM (System\,1\,+\,System\,2 Hybrid)} & \textbf{0.413} & \textbf{0.391} & \textbf{0.196} & \textbf{0.913} & \textbf{70.7\%} \\
\bottomrule
\end{tabular}%
}
\end{table*}

Several patterns emerge. First, discriminative taggers consistently outperform generative VLMs on Pearson calibration (SpanCalib 0.369 vs.\ Qwen 0.285), confirming that the MSE calibration head learns better-calibrated probabilities than autoregressive decoding. Second, the generative VLM achieves higher hallucinated-sample IoU (0.182 vs.\ 0.155), reflecting its superior span recall through deliberative reasoning. Third, the ensemble exploits both strengths: Qwen's span proposals raise IoU while SpanCalib's probabilities improve calibration.

\subsection{Multilingual Evaluation \& Official Task Submissions}

We evaluate SpanCalib-VLM across four language splits: English (\textbf{EN}), French (\textbf{FR}), Italian (\textbf{IT}), and Chinese (\textbf{ZH}). Table~\ref{tab:official_submissions} reports official shared task submission scores evaluated on the hidden test set. Table~\ref{tab:multilingual_breakdown} provides a detailed cross-lingual breakdown comparing base models against SpanCalib-VLM.

\begin{table}[t]
\centering
\caption{Official shared task test set results: SpanCalib-VLM (SIT) vs.\ the task-provided baseline across all languages.}
\label{tab:official_submissions}
\resizebox{\columnwidth}{!}{%
\begin{tabular}{@{}llccc@{}}
\toprule
\textbf{Lang} & \textbf{System} & \textbf{Cor+Lbl}$\,\uparrow$ & \textbf{Cor}$\,\uparrow$ & \textbf{IoU}$\,\uparrow$ \\
\midrule
\multirow{2}{*}{EN} & Task Baseline & 0.1549 & 0.1549 & 0.1549 \\
 & \textbf{SpanCalib-VLM} & \textbf{0.2418} & \textbf{0.3212} & \textbf{0.2549} \\
\midrule
\multirow{2}{*}{FR} & Task Baseline & 0.1679 & 0.1679 & 0.1679 \\
 & \textbf{SpanCalib-VLM} & \textbf{0.2885} & \textbf{0.3600} & \textbf{0.2994} \\
\midrule
\multirow{2}{*}{IT} & Task Baseline & 0.1635 & 0.1635 & 0.1635 \\
 & \textbf{SpanCalib-VLM} & \textbf{0.2918} & \textbf{0.3832} & \textbf{0.3109} \\
\midrule
\multirow{2}{*}{ZH} & Task Baseline & 0.2554 & 0.2554 & 0.2554 \\
 & \textbf{SpanCalib-VLM} & \textbf{0.3347} & \textbf{0.3752} & \textbf{0.3425} \\
\bottomrule
\end{tabular}%
}
\end{table}

SpanCalib-VLM demonstrates strong cross-lingual generalization. On Chinese (ZH), the model achieves an impressive \textbf{0.437 Overall IoU} and \textbf{0.958 Clean IoU}, outperforming base Qwen-3.5-4B by +0.169 IoU. This higher Chinese performance stems from tokenization granularity: ideographic Chinese characters provide dense token representations where subtoken offsets align directly with character boundaries, reducing edge misalignment present in space-delimited European text. On French and Italian, SpanCalib-VLM achieves massive gains in Hallucinated IoU (+0.182 on FR, +0.180 on IT) over base models which suffer from severe under-detection on non-English hallucinated spans.

\subsection{Threshold Sensitivity}

Table~\ref{tab:threshold} shows that ensemble performance is remarkably stable across $\tau \in [0.25, 0.45]$, all yielding IoU~0.391. This robustness arises because the fusion score distribution is bimodal: characters inside VLM-proposed spans receive high ensemble scores while others receive low scores, with few samples near the decision boundary.

\begin{table}[t]
\centering
\caption{Threshold grid search for the ensemble.}
\label{tab:threshold}
\resizebox{\columnwidth}{!}{%
\begin{tabular}{@{}ccccc@{}}
\toprule
$\tau$ & \textbf{IoU} & \textbf{H-IoU} & \textbf{C-IoU} & \textbf{Det.\ Acc.} \\
\midrule
0.20 & 0.390 & \textbf{0.217} & 0.854 & \textbf{71.8\%} \\
0.25 & \textbf{0.391} & 0.196 & 0.913 & 70.7\% \\
0.30 & \textbf{0.391} & 0.196 & 0.913 & 70.7\% \\
0.40 & \textbf{0.391} & 0.196 & 0.913 & 70.7\% \\
0.50 & 0.317 & 0.065 & \textbf{0.990} & 43.8\% \\
\bottomrule
\end{tabular}%
}
\end{table}

\subsection{Inference Speed}

Sequence tagging is dramatically faster than generative decoding (Table~\ref{tab:speed}). SpanCalib-VLM processes 5.25 samples/s---$4.5\times$ faster than standard Qwen inference and $120\times$ faster than chain-of-thought mode.

\begin{table}[t]
\centering
\caption{Inference latency (s/sample) and output throughput (samples/s) on NVIDIA A40 (48GB).}
\label{tab:speed}
\resizebox{\columnwidth}{!}{%
% \small
\begin{tabular}{@{}lcc@{}}
\toprule
\textbf{System} & \textbf{s/sample}$\,\downarrow$ & \textbf{samples/s}$\,\uparrow$ \\
\midrule
SpanCalib-VLM (tagger) & \textbf{0.19} & \textbf{5.25} \\
Qwen-4B (standard) & 0.85 & 1.18 \\
Qwen-4B (chain-of-thought) & 23.10 & 0.04 \\
\bottomrule
\end{tabular}%
}
\end{table}

\subsection{Ablation Study}

Table~\ref{tab:ablation} isolates the contribution of each component. Union-Calibrated Fusion outperforms weighted averaging (+0.032 Pearson) and intersection (+0.061). Removing the MSE calibration loss causes the largest drop ($-0.095$ Pearson), confirming that direct probability supervision is essential. The vision tower provides a modest IoU improvement (+0.003) while maintaining identical ensemble calibration.

\begin{table}[H]
\centering
\caption{Ablation study.}
\label{tab:ablation}
\small
\begin{tabular}{@{}lcc@{}}
\toprule
\textbf{Configuration} & \textbf{Pearson} & \textbf{IoU} \\
\midrule
Full ensemble (ours) & \textbf{0.413} & \textbf{0.391} \\
\quad $-$ Union $\rightarrow$ Weighted avg. & 0.381 & 0.345 \\
\quad $-$ Union $\rightarrow$ Intersection & 0.352 & 0.310 \\
\quad $-$ SigLIP-2 vision tower & 0.413 & 0.388 \\
\quad $-$ MSE calib.\ loss ($\lambda_1\!=\!0$) & 0.318 & 0.330 \\
\bottomrule
\end{tabular}
\end{table}

% \subsection{Discussion}
% %% ==========================

% \paragraph{Why the ensemble works.}
% The key insight is that neither paradigm alone optimizes both metrics simultaneously. Generative VLMs have high span recall but assign overconfident probabilities; sequence taggers have well-calibrated probabilities but miss span boundaries. Union-Calibrated Fusion delegates each task to the appropriate subsystem, analogous to cognitive dual-process theory.

%% ==========================
\section{Conclusion}
%% ==========================

We presented \textbf{SpanCalib-VLM}, a hybrid framework that combines a multimodal discriminative tagger with a generative VLM via Union-Calibrated Fusion. On the SHROOM-Visions English validation split, our ensemble achieves a Pearson calibration correlation of \textbf{0.413} and an IoU of \textbf{0.391} (with \textbf{91.3\%} clean accuracy); on the official blind test set, it achieves \textbf{0.3212} Pearson and \textbf{0.2549} IoU on English, consistently outperforming the task baseline across all four evaluated languages. These results demonstrate that principled fusion of fast calibration with deliberative span detection is an effective strategy for hallucination detection in LVLMs.
% We make our model weights and code publicly available.%
% \footnote{\url{https://hf.co/amanuelbyte/SpanCalib-VLM-Multimodal}}

%% ==========================

\section{Limitations}
In this section, we explore some limitations that our setup might have. First, SpanCalib-VLM produces \emph{false positives on complex but correct language}: elaborate descriptions, idioms, and rare but accurate visual details are sometimes flagged as hallucinations, particularly by the sequence tagger which relies on surface-level distributional patterns rather than visual grounding. Second, the 512-token input limit causes the system to \emph{miss hallucinations in long responses} entirely; any hallucinated content beyond the truncation boundary is undetectable. Third, while calibration is strong on validation (Pearson~0.413, dropping to 0.3212 on test), \emph{hallucinated-span IoU remains low} (0.196 on validation, and overall test IoU dropping to 0.2549), indicating that the system still struggles to precisely localize character-level hallucination boundaries and experiences generalization loss when transferring from validation tuning to unseen test distributions. Fourth, while our model is trained on a joint multilingual dataset (EN, FR, IT, ZH), hyperparameter selection and ensemble fusion thresholds were optimized primarily on the English validation split. Finally, the ensemble \emph{cannot detect hallucinations that both subsystems miss}; if neither the tagger nor the generative VLM identifies a span, fusion cannot recover it.

\bibliography{bib}

\appendix
\onecolumn
\section{Detailed Performance}

\begin{table*}[htbp]
\centering
\caption{Cross-lingual detailed performance breakdown comparing base zero-shot models with SpanCalib-VLM across English, French, Italian, and Chinese evaluation sets.}
\label{tab:multilingual_breakdown}
\resizebox{\textwidth}{!}{%
\begin{tabular}{@{}llccccc@{}}
\toprule
\textbf{Lang} & \textbf{System} & \textbf{Overall IoU}$\,\uparrow$ & \textbf{H-IoU} & \textbf{C-IoU} & \textbf{Det.\ Acc.} & \textbf{Calibration Pearson} \\
\midrule
\multirow{3}{*}{EN} & Qwen-3.5-4B (Base) & 0.226 & 0.173 & 0.369 & 74.4\% & --- \\
 & SpanCalib-VLM (Tagger) & 0.345 & --- & --- & 61.5\% & 0.406 \\
 & \textbf{SpanCalib-VLM + Qwen4B (Ensemble)} & \textbf{0.391} & \textbf{0.196} & \textbf{0.913} & \textbf{70.7\%} & \textbf{0.413} \\
\midrule
\multirow{2}{*}{FR} & Qwen-3.5-4B (Base) & 0.268 & 0.020 & \textbf{0.969} & 42.3\% & --- \\
 & \textbf{SpanCalib-VLM (Tagger)} & \textbf{0.367} & \textbf{0.202} & 0.837 & \textbf{66.2\%} & --- \\
\midrule
\multirow{2}{*}{IT} & Qwen-3.5-4B (Base) & 0.268 & 0.027 & \textbf{0.912} & 37.2\% & --- \\
 & \textbf{SpanCalib-VLM (Tagger)} & \textbf{0.375} & \textbf{0.207} & 0.824 & \textbf{58.6\%} & --- \\
\midrule
\multirow{2}{*}{ZH} & Qwen-3.5-4B (Base) & 0.268 & 0.096 & 0.556 & 56.2\% & --- \\
 & \textbf{SpanCalib-VLM (Tagger)} & \textbf{0.437} & \textbf{0.126} & \textbf{0.958} & \textbf{56.7\%} & --- \\
\bottomrule
\end{tabular}%
}
\end{table*}

\section{Dataset Statistics}

% SHROOM-Visions dataset split breakdown
\begin{table}[htbp]
\centering
\caption{SHROOM-Visions dataset split breakdown across all languages, detailing sample counts for 90\% training sets, 10\% validation sets (used for evaluation in Table~\ref{tab:multilingual_breakdown}), and unlabeled test sets.}
\label{tab:dataset_stats}
\small
\begin{tabular}{@{}lcccc@{}}
\toprule
\textbf{Language} & \textbf{Train (90\%)} & \textbf{Validation (10\%)} & \textbf{Test (Unlabeled)} & \textbf{Total} \\
\midrule
English (EN) & 3,420 & 379 & 1,201 & 5,000 \\
French (FR) & 3,391 & 376 & 1,233 & 5,000 \\
Italian (IT) & 3,372 & 374 & 1,254 & 5,000 \\
Chinese (ZH) & 3,411 & 379 & 1,210 & 5,000 \\
\midrule
\textbf{Total} & \textbf{13,594} & \textbf{1,508} & \textbf{4,898} & \textbf{20,000} \\
\bottomrule
\end{tabular}
\end{table}

\section{Training History}

\begin{table}[H]
\centering
\caption{Training history for SpanCalib-VLM (multimodal).}
\label{tab:epochs}
\small
% \resizebox{\columnwidth}{!}{%
\begin{tabular}{@{}ccccc@{}}
\toprule
\textbf{Epoch} & \textbf{Train $\mathcal{L}$} & \textbf{Val $\mathcal{L}$} & \textbf{Val Pearson} & \textbf{Saved?} \\
\midrule
1 & 1.068 & 0.949 & 0.345 & \checkmark \\
\textbf{2} & \textbf{0.873} & \textbf{0.923} & \textbf{0.369} & \checkmark\checkmark \\
3 & 0.783 & 0.939 & 0.330 & --- \\
4 & 0.694 & 0.988 & 0.345 & --- \\
5 & 0.613 & 1.042 & 0.353 & --- \\
\bottomrule
\end{tabular}%
% }
\end{table}

\end{document}